\documentclass[]{spie}  %>>> use for US letter paper
\usepackage{amsmath,amsfonts,amssymb}
\usepackage{graphicx}
\usepackage{multirow}
\usepackage{adjustbox}
\usepackage{booktabs}

\usepackage[colorlinks=true, allcolors=blue]{hyperref}

\title{Benchmarking Multi-Organ Segmentation Tools for Multi-Parametric T1-weighted Abdominal MRI}

\author{Nicole Tran, Anisa Prasad, Yan Zhuang, Tejas Sudharshan Mathai, Boah Kim, \\ Sydney Lewis, Pritam Mukherjee, Jianfei Liu, Ronald M. Summers}
\affil{Imaging Biomarkers and Computer-Aided Diagnosis Laboratory, Radiology and Imaging Sciences, National Institutes of Health Clinical Center, Bethesda, USA}

\authorinfo{Send correspondence to T.S.M.: tejas dot mathai at nih dot gov}

\begin{document} 
\maketitle

\begin{abstract}

The segmentation of multiple organs in multi-parametric MRI studies is critical for many applications in radiology, such as correlating imaging biomarkers with disease status (e.g., cirrhosis, diabetes). Recently, three publicly available tools, such as MRSegmentator (MRSeg), TotalSegmentator MRI (TS), and TotalVibeSegmentator (VIBE), have been proposed for multi-organ segmentation in MRI. However, the performance of these tools on specific MRI sequence types has not yet been quantified. In this work, a subset of 40 volumes from the public Duke Liver Dataset was curated. The curated dataset contained 10 volumes each from the pre-contrast fat saturated T1, arterial T1w, venous T1w, and delayed T1w phases, respectively. Ten abdominal structures were manually annotated in these volumes. Next, the performance of the three public tools was benchmarked on this curated dataset. The results indicated that MRSeg obtained a Dice score of 80.7 $\pm$ 18.6 and Hausdorff Distance (HD) error of 8.9 $\pm$ 10.4 mm. It fared the best ($p < .05$) across the different sequence types in contrast to TS and VIBE. 

% There is a need for a robust multi-organ segmentation tool to correlate MRI-based organ biomarkers with disease statuses, such as in diabetes or liver cirrhosis. Currently, 3 such tools have been published: TotalSegmentator MRI (TS), MRSegmentator (MRSeg), and TotalVibeSegmentator (VIBE). These tools must function effectively on all MRI sequence types. To quantify and compare the performance of these tools on different sequence types, we curate a database of 10 abdominal structures annotated in 4 T1 MRI sequences.

\end{abstract}

\keywords{MRI, Multi-Parametric, T1-weighted, Segmentation, Abdomen}

% ================================================
\section{Introduction}
\label{sec_intro}  % \label{} allows reference to this section
% ================================================

Magnetic Resonance Imaging (MRI) is a widely used imaging modality that is useful for many applications, such as early detection and diagnosis of diseases \cite{macdonald2023duke, hussain2022modern,branca2010molecular, eustace2004whole}, radiotherapy planning and guidance~\cite{keall2022integrated, otazo2021mri,dirix2014value}, and many others \cite{zaffina2022body, huber2020mri, nyholm2014counterpoint, yu2016emergency, hosny2018artificial}. Segmentation of various abdominal structures (e.g., liver, lungs, and kidneys) is a necessity for several applications, but obtaining them can be challenging due to a dearth of publicly available datasets with high quality annotations that can be used to train a segmentation model. In fact, obtaining such labels is time-consuming and labor-intensive, and therefore infeasible for a clinician to perform during a busy clinical day \cite{hosny2018artificial,greenspan2016guest,zhu2023utilizing}.

To obtain organ segmentations without any clinician intervention, numerous studies have explored organ segmentation in MRI for the spine~\cite{zhang2020sau}, chest~\cite{weng2021deep}, abdomen~\cite{ji2022amos,macdonald2023duke}, pelvis~\cite{nyholm2018mr,zhuang2024segmentation}, and knee~\cite{panfilov2019improving}. Previously, multi-organ segmentation in MRI lagged significantly behind its CT counterpart \cite{wasserthal2023totalsegmentator}. However, recent advancements in multi-organ and structure segmentation \cite{Zhuang2024,Hantze2024mr,Dantonoli2024ts,Graf2024vibe} have closed this gap. These models have been trained on heterogeneous datasets with diverse patients, different exam protocols, and various sequence types. Moreover, these prior works have only been validated on the external AMOS22 testing dataset \cite{ji2022amos}. Unfortunately, information corresponding to patient demographics and data acquisition parameters were not made publicly available with this dataset. Additionally, annotations were only provided for 13 key abdominal organs across 60 patients. Therefore, the bias of these tools towards one or more MRI sequence types is presently not known. {A tool that allows for analysis of all sequence types is crucial: pre-contrast MRI establishes the baseline tissue characteristics, the arterial and venous phases highlight vascular structures, and the delayed phase reveals contrast retention patterns, aiding tissue differentiation.~\cite{murphy2021}}

% This challenge is amplified in the case of the abdomen, which is an anatomically complex area with several subtle structures. As a result, there is a noticeable lack of large-scale, high-quality MRI datasets with voxel-level annotation, detailed information on MRI acquisition, and transparent data on patient demographics. 

In this study, we benchmark the performance of three publicly available multi-organ MRI segmentation tools against each other and across sequence types. For this purpose, a multi-parametric abdominal T1 MRI dataset was curated from the public Duke Liver Dataset \cite{macdonald2023duke}. The data subset contained 10 volumes each from pre-contrast T1-weighted (T1w PRE), contrast-enhanced T1-weighted MRI in the arterial (T1w ART), portal venous (T1w VEN), and delayed (T1w DEL) phases, thereby totalling 2838 2D slices from 34 unique patients (40 volumes). Voxel-level annotations for 10 structures across various regions in the abdomen were obtained. Next, the performance of the three tools were evaluated for their capability to segment structures in this curated T1 MRI dataset. The robustness of these tools was tested on a dataset that was entirely out of the training distribution of each tool.

% 700 in arterial
% 698 in delayed
% 732 in precontrast
% 708 in venous
% total 2838 slices

% There is a need for a robust multi-organ segmentation tool to correlate MRI-based organ biomarkers with disease statuses, such as in diabetes or liver cirrhosis. These tools would be useful for applications including diagnosing and tracking disease progression, opportunistic disease screening \cite{Zaffina2022}, and pre- and post- surgery imaging. While the segmentation could be done by human radiologists, manual annotation is an extremely time consuming process that makes automated MRI segmentation a much less costly and less subjective alternative.

% At the time of this project, 

% It is necessary for these tools to be able to work well on all of the MRI sequence types, regardless of the type coming in- some, like TS, claim to have a sequence-independent tool. Although this may be true, it is important to examine whether a tool works better on one sequence over another, especially in relation to training data. Tools such as MR used more CT scans than anything else, TS used mainly T1 sequences, and Vibe used T2 sequences in their training data - examining these in relation to their performance on different sequences will inform whether it is worth investing into even distributions of sequence types in future training data.

% In this paper, we use the Duke Liver Dataset to compare performance of the three tools against each other, and performance of each tool on different sequence types.

%%%%%%%%%%%%%%%%%%%%%%%%%%%%%%%%%%%%%%%%%%%%%%%%%%%%%%%%%%%%%%%%%%
\begin{figure*}[!t]
% \begin{minipage}[b]{0.95\columnwidth}
%   \centering
%   \centerline{}
% %  \vspace{2.0cm}
%   \centerline{(a) Training Pipeline 1}\medskip
% \end{minipage}
\centering
\includegraphics[width=1\textwidth]{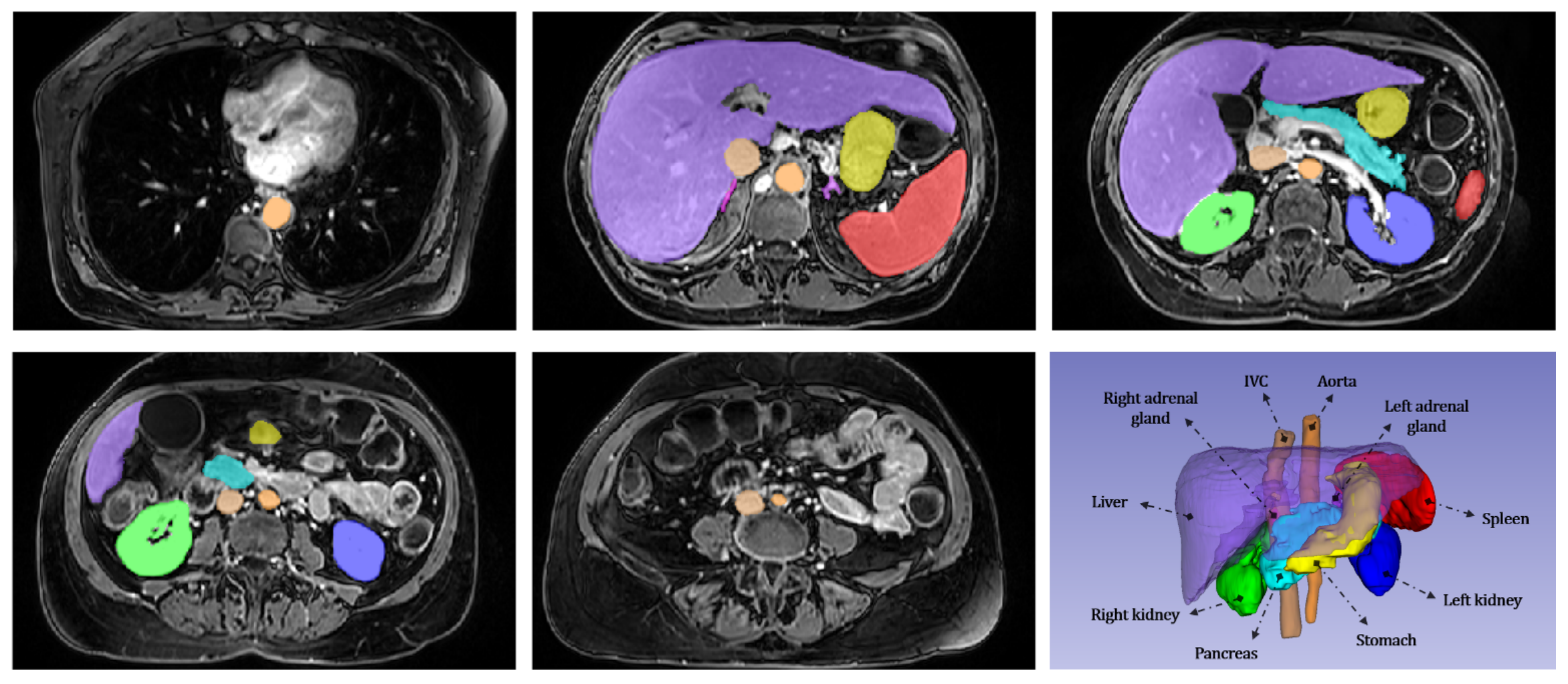}
\caption{We curated a subset of the Duke Liver dataset consisting of 40 volumes, 10 each from pre-contrast T1, arterial T1w, venous T1w, and delayed T1w series. 10 common abdominal organs (bottom right) were manually segmented in these volumes and verified by a senior board-certified radiologist. Examples of the manual segmentations for these structures at different slices (from superior to inferior) in one scan are shown.}
\label{fig:fig_money}
\end{figure*}
%%%%%%%%%%%%%%%%%%%%%%%%%%%%%%%%%%%%%%%%%%%%%%%%%%%%%%%%%%%%%%%%%%

% ================================================
\section{METHODS}
\label{sec_methods}
% ================================================

\textbf{Patient Sample.} The Duke Liver Dataset \cite{macdonald2023duke} was used in this work. It consisted of 2146 MRI sequences from 105 patients (76 men, 29 women; age range, 30–80 years). The patients underwent contrast-enhanced MRI imaging at three centers with 87 patients showing imaging findings of cirrhosis. The MRI studies were obtained with Siemens (n = 96) and GE (n = 9) scanners and the magnetic field strengths of these scanners varied (54 with 1.5T, 51 with 3T). A total of 17 different MRI sequence types (multi-planar, multi-phase) were available in this dataset. 

\noindent
\textbf{T1 MRI Benchmark Dataset Creation.} Only the axial T1-weighted (T1w) sequences from the Duke Liver Dataset were considered. The dataset had only 2 T2w sequences (T2w and T2 fat suppressed), whereas there were 6 T1 sequences to assess the performance of public MRI organ segmentation tools. Following the descriptions outlined in Zhu et al. \cite{Zhu2022}, the 6 different phases were consolidated into 4 coarse groups that included: (1) pre-contrast fat suppressed T1w, (2) dynamic arterial T1w (combination of early, mid, and late arterial), (3) dynamic venous T1w, and (4) dynamic delayed T1w. Ten volumes were randomly selected from each group to be included in the benchmark dataset, which resulted in a total of 40 T1w volumes. {Out of the T1w volumes, 35 came from unique patients, and some patients had been imaged multiple times during different visits.}

In these volumes, 10 structures were manually labeled by a grader (2 years of experience) and included: (1) spleen, (2) left kidney, (3) right kidney, (4) stomach, (5) aorta, (6) inferior vena cava, (7) pancreas, (8) left adrenal gland, (9) right adrenal gland, and (10) liver. Labeling 10 structures in 1 volume took $\sim$5 hours, and a total of $\sim$215 hours were required to annotate all 10 structures in 40 volumes. This highlights the cumbersome nature of the annotations, which were fully reviewed by a senior board-certified radiologist (30+ years of experience).

\noindent
\textbf{Public MRI Multi-Organ Segmenters.} Presently, three multi-organ MRI segmentation tools are publicly available. These include: MRSegmentator (MRSeg) \cite{Hantze2024mr}, TotalSegmentator MRI (TS) \cite{Dantonoli2024ts}, and TotalVibeSegmentator (VIBE) \cite{Graf2024vibe}. MRSeg, TS, and VIBE were evaluated on the 40 T1 volumes in our curated dataset, and segmented 40, 59, and 71 structures, respectively. A summary of the dataset characteristics that each model was trained and tested on (including external validations) is presented in the Appendix.  

% The segmentation masks from each individual tool were tested against the ground truth on a organ by organ basis for each volume using DSC scores and Hausdorff distances. For cases where patients were missing and organ or an organ could not be detected (as was the case for one kidney and one adrenal gland in the precontrast images), the scores for that organ for that volume were excluded from the comparison calculations and figures.

\noindent
\textbf{Statistical Analysis.} The segmentation performance was quantitatively measured using Dice similarity coefficient (DSC) and Hausdorff Distance (HD) error. A Friedman test was performed to statistically compare the performance of the three segmentation tools for each sequence type, and a post-hoc Nemenyi test determined any specific differences between the approaches. 

% ================================================
\section{Results}
\label{sec_results}
% ================================================

\begin{figure*}[!t]
% \begin{minipage}[b]{0.95\columnwidth}
%   \centering
%   \centerline{}
% %  \vspace{2.0cm}
%   \centerline
% \end{minipage}
\centering
\includegraphics[width=0.7\textwidth]{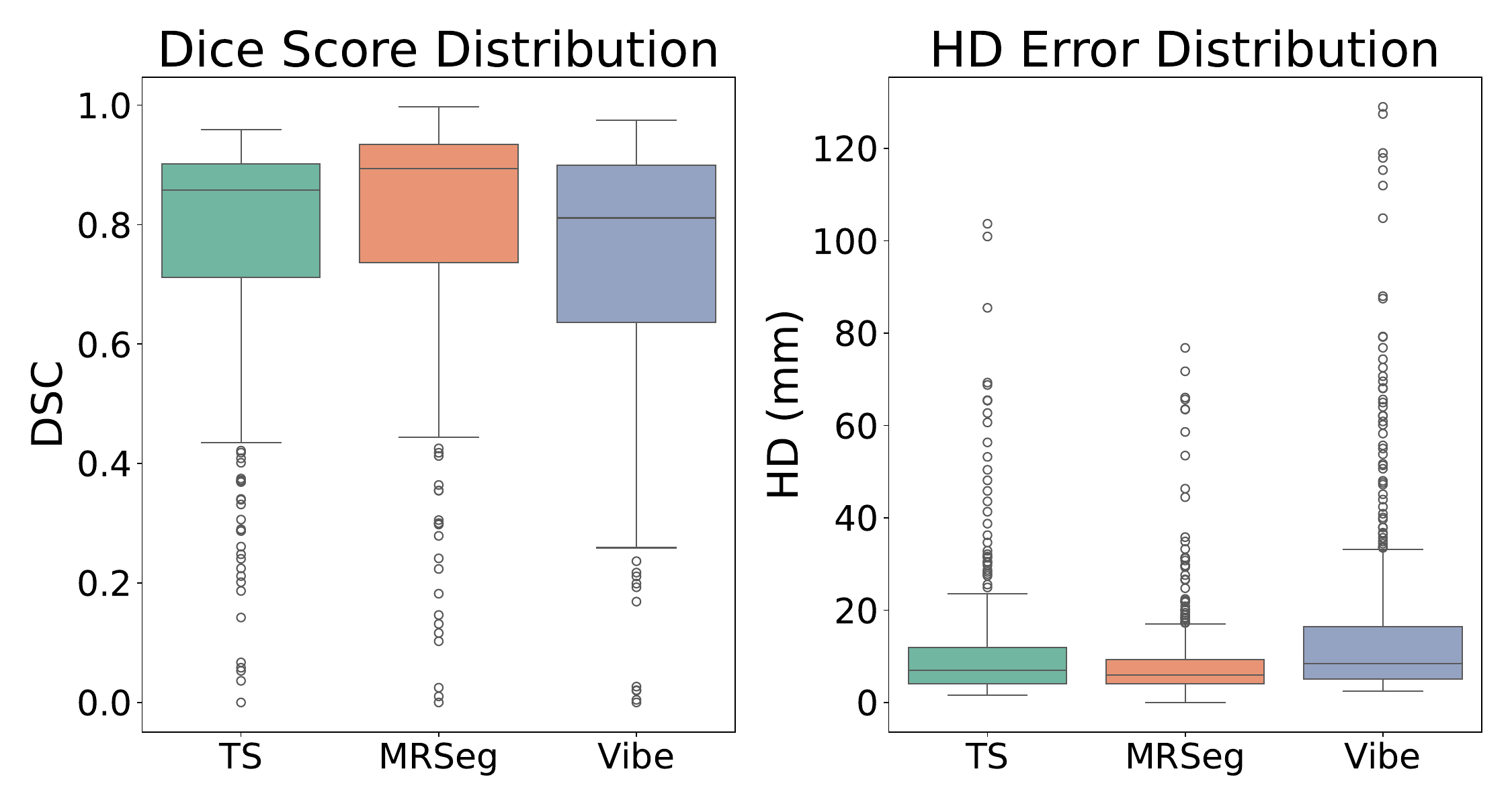}
\caption{Comparison of the DSC and Hausdorff Distance (HD) errors across all 10 structures in 40 volumes for the different multi-organ MRI segmenters.}
\label{fig_overall_results}
\end{figure*}

The Dice scores and HD errors for the three segmentation tools across each sequence type are shown in Table \ref{tab_overall_results}. Fig. \ref{fig_overall_results} shows the distribution of DSC and HD errors for each tool across the 40 volumes in the dataset. Overall, MRSeg obtained the highest Dice score of 80.7 $\pm$ 18.6 and lowest HD error of 8.9 $\pm$ 10.4 mm across all the sequence types. Supplemental Tables \ref{tab_dsc_friedman} and \ref{tab_hd_friedman} describe the p-values from the statistical tests. Across all sequences, a difference in segmentation performance (both DSC and HD) was observed between the three tools ($p < .001$). In terms of Dice score, differences were seen between model pairs ($p < .05$) for all sequences, except that there was no difference in performance between TS and VIBE ($p = 0.1$) for the T1w arterial sequence. With respect to HD errors, no difference in performance was seen between TS vs. VIBE for the pre-contrast T1 series ($p = .104$), and TS vs. MRSeg ($p = .073$) and TS vs. VIBE ($p = .093$) for the arterial series, respectively. 

\begin{figure*}[!t]
% \begin{minipage}[b]{0.95\columnwidth}
%   \centering
%   \centerline{}
% %  \vspace{2.0cm}
%   \centerline
% \end{minipage}
\centering
\includegraphics[width=0.9\textwidth]{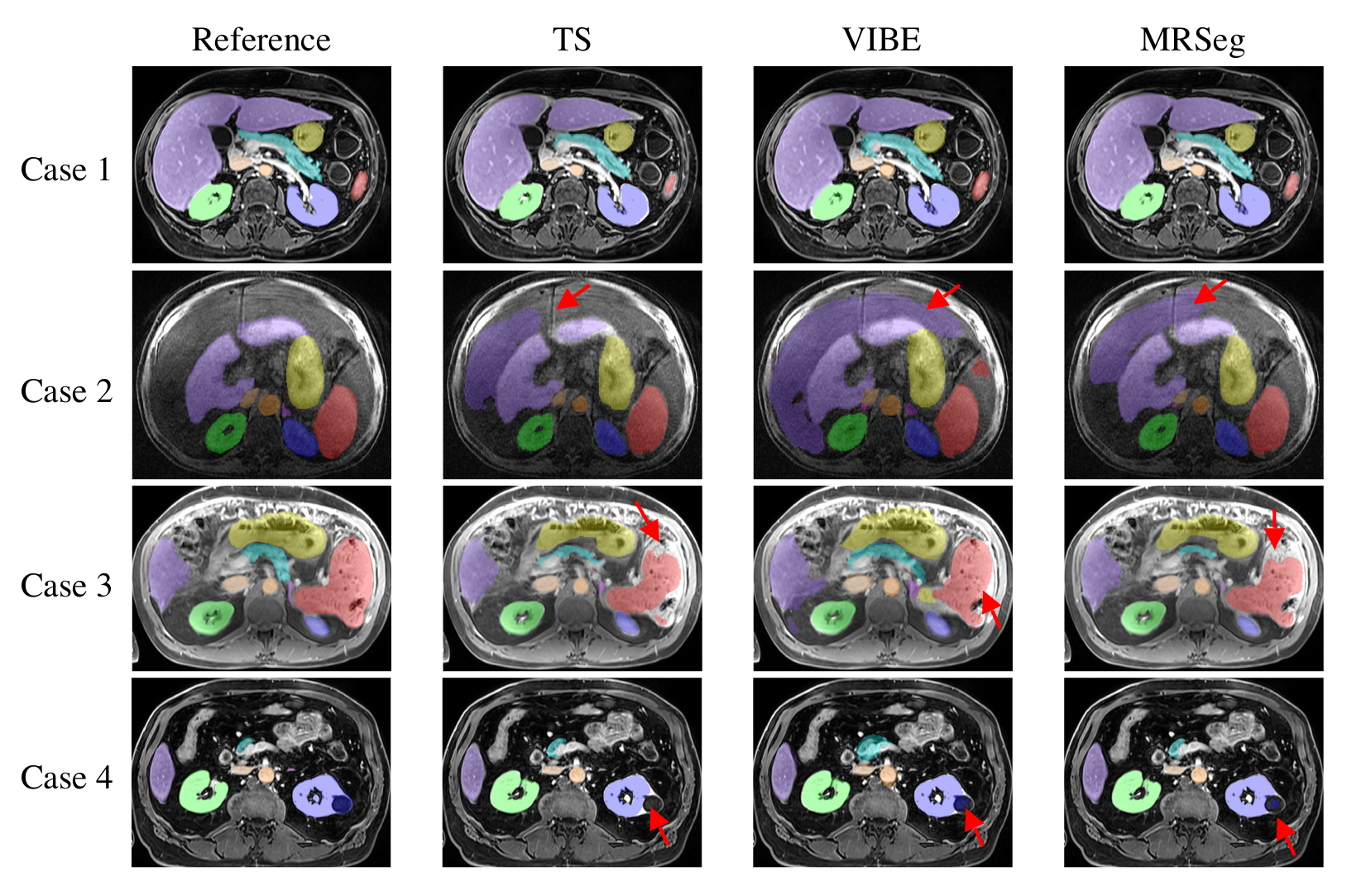}
\caption{Comparison of multi-organ segmentations by TS, VIBE, and MRSeg for four different patients containing various disease conditions. Case 1 shows a normal patient with no disease. Case 2 shows a patient with liver cirrhosis. Note the over-segmentation of the liver into adjacent ascites (fluid region, red arrows). Case 3 shows a patient with multiple splenic lesions (red arrows). Case 4 shows a patient with a lesion in the left kidney.}
\label{fig_results_qualitative}
\end{figure*}

Fig. \ref{fig_results_qualitative} visually illustrates the segmentation results by the three segmentation tools for a few cases. All the tools struggled with the pathologies present in the Duke Liver Dataset, such as cirrhosis {or the presence of kidney lesions, tending to undersegment in the case of lesions and oversegment in the case of cirrhosis}. The performance of the three tools on each of the 10 structures are shown in Supplemental Figs. \ref{fig_DSC_largeOrgans} to \ref{fig_hd_smallOrgans}. MRSeg consistently obtained the highest DSC and lowest HD errors for large organs (liver, spleen, stomach), medium-sized organs (kidneys and pancreas), and small organs (adrenal glands, aorta and inferior vena cava). Notably, MRSeg segmented the pancreas and the aorta better than TS and VIBE. VIBE had the highest HD errors across all structures; the error was greatest mainly for the stomach, aorta, and pancreas.  

All the tools over-segmented the liver and encroached into the adjacent Ascites (fluid buildup around the liver) as seen in Fig. \ref{fig_results_qualitative}. Notably, they under-segmented the pancreas and the adrenal glands, and did not segment lesions and cysts if they were present in certain organs, such as the spleen and kidneys. It is important to note that there were missing organs in two pre-contrast series; the left adrenal gland was missing from one series, while the right kidney was removed from another pre-contrast series. These missing organs were accounted for and the presented results are shown for those organs that were available. TS and VIBE had false positive segmentations for these missing structures as shown in Supplemental Fig. \ref{fig_falsePositiveSegmentations}.

\begin{table}[!t]
\centering
\caption{DSC (\%) and Hausdorff Distance (mm) errors for each multi-organ MRI segmenter are shown across all T1 sequences. Bold font indicates best results.}
\smallskip
\begin{adjustbox}{max width=0.95\textwidth}
\begin{tabular}{@{} l *{6}{c} @{}}
  \toprule
  Dataset &
    \multicolumn{3}{c}{DSC (\%) $\uparrow$} &
    \multicolumn{3}{c}{HD (mm) $\downarrow$} \\
  \cmidrule(lr){2-4} \cmidrule(l){5-7}
  & TS & MRSeg & VIBE & TS & MRSeg & VIBE \\
  \midrule
  Pre-Contrast & 76.5 $\pm$ 17.9 & \textbf{79.8 $\pm$ 17.2} & 77.9 $\pm$ 17.3 & 10.4 $\pm$ 11.5 & \textbf{9.1 $\pm$ 9.9} & 15.0 $\pm$ 18.3 \\
  Arterial & 76.0 $\pm$ 17.3 & \textbf{78.3 $\pm$ 18.3} & 72.7 $\pm$ 18.9 & 12.3 $\pm$ 13.6 & \textbf{9.9 $\pm$ 10.0} & 16.9 $\pm$ 19.9 \\
  Venous & 80.5 $\pm$ 17.1 & \textbf{84.1 $\pm$ 16.7} & 73.9 $\pm$ 20.4 & 10.3 $\pm$ 15.3 & \textbf{6.8 $\pm$ 7.6} & 18.5 $\pm$ 27.1 \\
  Delayed & 77.7 $\pm$ 21.4 & \textbf{80.7 $\pm$ 21.3} & 72.5 $\pm$ 23.3 & 10.2 $\pm$ 11.5 & \textbf{9.9 $\pm$ 13.1} & 15.1 $\pm$ 17.3 \\
  All & 77.7 $\pm$ 18.6 & \textbf{80.7 $\pm$ 18.6} & 74.3 $\pm$ 20.2 & 10.8 $\pm$ 13.1 & \textbf{8.9 $\pm$ 10.4} & 16.4 $\pm$ 20.1 \\
  \bottomrule
\end{tabular}
\end{adjustbox}
\label{tab_overall_results}
\end{table}
%%%%%%%%%%%%%%%%%%%%%%%%%%%%%%%%%%%%%%%%%%%%%%%%%%%%%%%%%%%%%%%%%%

%%%%%%%%%%%%%%%%%%%%%%%%%%%%%%%%%%%%%%%%%%%%%%%%%%%%%%%%%%%%%%%%%%
%% excluding scores/distances for one left adrenal gland and one right kidney due to the organ/structure not being present in the scan.
%% MRSeg performed best (corrected p-value $>$  0.001)
%%%%%%%%%%%%%%%%%%%%%%%%%%%%%%%%%%%%%%%%%%%%%%%%%%%%%%%%%%%%%%%%%%

% ================================================
\section{Discussion}
\label{sec_discussion}
% ================================================

All the compared segmentation tools used the nnUNet architecture for training their model. The superior performance of the MRSeg tool can be attributed to the underlying training dataset \cite{Hantze2024mr}, which consisted of 1200 Dixon MRI studies from 50 patients in the UK Biobank, 221 MRI sequences from their internal German institution (an equal distribution of T1, T2 and T1 fat saturated MRI series), and the entirety of the TotalSegmentator CT dataset (1228 series). {All tools used an iterative learning process to generate the annotations for their training datasets. VIBE was the only tool to train on exclusively MRI volumes. \cite{Graf2024vibe} Both MRSeg and TS used CT volumes in their training data \cite{Hantze2024mr, wasserthal2023totalsegmentator}. However, TS was the only tool to not use the CT-based TotalSegmentator for segmentation of any new volumes.} MRSeg leveraged several different sources of MRI and CT data for training, and posted the best performance on our curated dataset of only T1 sequences. From the publication of this tool, it is known that the tool fared the best on T1 opposed phase series. From the T1 sequences evaluated in this work, the tool performed well on the T1w venous and T1w delayed sequences, respectively. 

The failure cases with MRSeg are also known issues \cite{Hantze2024mr} because it cannot segment small organs well, such as adrenal glands, resulting in low dice scores. Similarly, the under-segmentation of organs containing lesions were due to the heterogeneous appearance and irregular borders of the lesions compared to the parenchyma. Interestingly, TS was unable to attain the same level of performance as MRSeg despite being trained on a variety of multi-parametric MRI and CT studies. This shows that generalized tools for multi-organ segmentation, which can be versatile and broadly applicable for many applications, sometimes do not obtain high segmentation accuracy compared to tools that are specifically tailored towards abdominal organ segmentation \cite{Hou2024,Zhuang2024}. {Similar results were found in TS's own evaluation against MRSeg, where TS fell short in the abdominal region but outperformed MRSeg for other structures. \cite{wasserthal2023totalsegmentator}} {It also is interesting to note that using CT volumes in the training data results in an overall better performance, as seen from MRSeg and TS outperforming VIBE. This is not unexpected, as seen in TS's own ablation studies, but should be considered for future training dataset curation. \cite{wasserthal2023totalsegmentator}}

In summary, three publicly available multi-organ MRI segmentation tools were benchmarked on a curated dataset of T1 sequences. The effect of the sequence type on a tool's segmentation performance was quantified. MRSegmentator fared the best for the different T1 sequence types for axial abdominal images, followed by TotalSegmentator MRI. 

%%%%%%%%%%%%%%%%%%%%%%%%%%%%%%%%%%%%%%%%%%%%%%%%%%%%%%%%%%%%%%%%%%
%%%%%%%%%%%%%%%%%%%%%%%%%%%%%%%%%%%%%%%%%%%%%%%%%%%%%%%%%%%%%%%%%%

% It's clear from both sequence by sequence and overall comparison that MRSeg had overall better performance than the other two tools, followed by TS. VIBE did worse overall, maybe because it was trained on breath-hold images, and the Duke Liver Dataset does not specify that it includes that type of scan. MRSeg also had the highest number of scans included in training data, including the entirety of the CT dataset from the original TotalSegmentator tool, and included and abdominal in-house dataset. Given that the images from the Duke Liver Dataset were also abdominal, this could have offered it an advantage. 

% However, we also note that the TS tool performed comparably to MRSeg on most sequence types, even if not as notably superior to VIBE when compared to MRSeg. In addition, no tool was overwhelmingly good or bad at any one T1w sequence type.

% Lastly, we also see that all three tools struggle to properly segment patients with pathologies, so we suggest adding more scans to training data including common pathologies.

% In this work we show that of the published MRI segmentation models, . No tool performed exceedingly well on any one sequence type, and all tools struggled with pathologies.

\clearpage

% ================================================
\section{Acknowledgements}
% ================================================

This work was supported by the Intramural Research Program of the National Institutes of Health, Clinical Center and used the computational resources of the NIH HPC Biowulf cluster.

% ================================================
\bibliography{references} % bibliography data in report.bib
% ================================================
% References
\bibliographystyle{spiebib} % makes bibtex use spiebib.bst

\clearpage

% ================================================
\section{Appendix}
% ================================================

\subsection{MRI multi-organ segmenters}

A summary of the three multi-organ segmentation tools for multi-parametric MRI sequences is described below. 

MRSegmentator (MRSeg) \cite{Hantze2024mr} was trained on 1,200 UK Biobank Dixon MRI exams (50 patients), 221 MRI sequences from an internal German dataset (177 patients with approximately equal distribution of T1, T2, and T1w fat saturated series), and the entire public TotalSegmentator CT dataset (1228 series). MRSeg segmented 40 structures, and obtained an average DSC of 0.85 $\pm$ 0.13 on the NAKO dataset and 0.79 $\pm$ 0.11 on the public AMOS22 dataset.

TotalSegmentator MRI (TS) \cite{Dantonoli2024ts} was trained on multi-parameteric MRI studies from 251 patients (147 men, 104 women, median age 60, age IQR: 47, 71) who were imaged at the University Hospital Basel. Additionally, 47 MRI images from the Imaging Data Commons as well as 227 CT series (135 patients, 74 men, 61 women, 97 unknown, median age 69, age IQR: 61, 77) from the TotalSegmentator CT dataset were used. TS segmented 59 structures and obtained an average Dice score of 0.824 (CI: 0.801, 0.842) on their internal test set (30 MRI volumes) and 0.801 (CI: 0.780, 0.824) on the public AMOS22 dataset. 

TotalVibeSegmentator (VIBE) \cite{Graf2024vibe} was trained on volumetric interpolated breath-hold examinations that used a two-point Dixon sequence to separate water and fat in MRI sequences. The training dataset contained full torso VIBE images (excluding head, and parts of arms and legs) from the NAKO (85 patients) and the UK Biobank (16 patients). VIBE segmented $>$71 labels in a held-out internal test set (12 patients) with an average DSC of 0.89 $\pm$ 0.07.

\subsection{Results}

%%%%%%%%%%%%%%%%%%%%%%%%%%%%%%%%%%%%%%%%%%%%%%%%%%%%%%%%%%%%%%%%%%
%% new table 1a
\begin{table}[!htb]
\centering
\caption{Statistical comparison (p-values) of the Dice scores from different segmenters (TS, MRSeg, VIBE) across the various sequence types. A p-value $<$ .05 indicated statistical significance.}
\smallskip
\begin{adjustbox}{max width=0.95\textwidth}
\begin{tabular}{@{} l *{4}{c} @{}}
  \toprule
  Sequence & Friedman p-value
  & TS vs. MRSeg & MRSeg vs. VIBE & TS vs. VIBE \\
  \midrule
  All & $< $ 0.001 & 0.001 & 0.001 & 0.001 \\
  Pre-Contrast & $< $ 0.001 & 0.001 & $< $ 0.001 & $< $ 0.001 \\
  Arterial & $< $ 0.001 & 0.020 & 0.001 & 0.100 \\
  Delayed & $< $ 0.001 & 0.001 & 0.001 & 0.001 \\
  Venous & $< $ 0.001 & 0.001 & 0.001 & 0.001 \\
  \bottomrule
\end{tabular}
\end{adjustbox}
\label{tab_dsc_friedman}
\end{table}
%%%%%%%%%%%%%%%%%%%%%%%%%%%%%%%%%%%%%%%%%%%%%%%%%%%%%%%%%%%%%%%%%%

%%%%%%%%%%%%%%%%%%%%%%%%%%%%%%%%%%%%%%%%%%%%%%%%%%%%%%%%%%%%%%%%%%
%% new table 1
\begin{table}[!htb]
\centering
\caption{Statistical comparison (p-values) of the Hausdorff Distance (HD) errors by different segmenters (TS, MRSeg, VIBE) across the various sequence types. A p-value $<$ .05 indicated statistical significance. }
\smallskip
\begin{adjustbox}{max width=0.95\textwidth}
\begin{tabular}{@{} l *{4}{c} @{}}
  \toprule
  Sequence & Friedman p-value
  & TS vs. MRSeg & MRSeg vs. VIBE & TS vs. VIBE \\
  \midrule
  All & $< $ 0.001 & 0.001 & 0.001 & 0.001 \\
  Pre-Contrast & $< $ 0.001 & 0.048 & 0.001 & 0.104 \\
  Arterial & $< $ 0.001 & 0.073 & 0.001 & 0.093 \\
  Delayed & $< $ 0.001 & 0.005 & 0.001 & 0.005 \\
  Venous & $< $ 0.001 & 0.036 & 0.001 & 0.001\\
  \bottomrule
\end{tabular}
\end{adjustbox}
\label{tab_hd_friedman}
\end{table}
%%%%%%%%%%%%%%%%%%%%%%%%%%%%%%%%%%%%%%%%%%%%%%%%%%%%%%%%%%%%%%%%%%

%%%%%%%%%%%%%%%%%%%%%%%%%%%%%%%%%%%%%%%%%%%%%%%%%%%%%%%%%%%%%%%%%%
\begin{figure*}[!htb]
% \begin{minipage}[b]{0.95\columnwidth}
%   \centering
%   \centerline{}
% %  \vspace{2.0cm}
%   \centerline
% \end{minipage}
\centering
\includegraphics[width=0.85\textwidth]{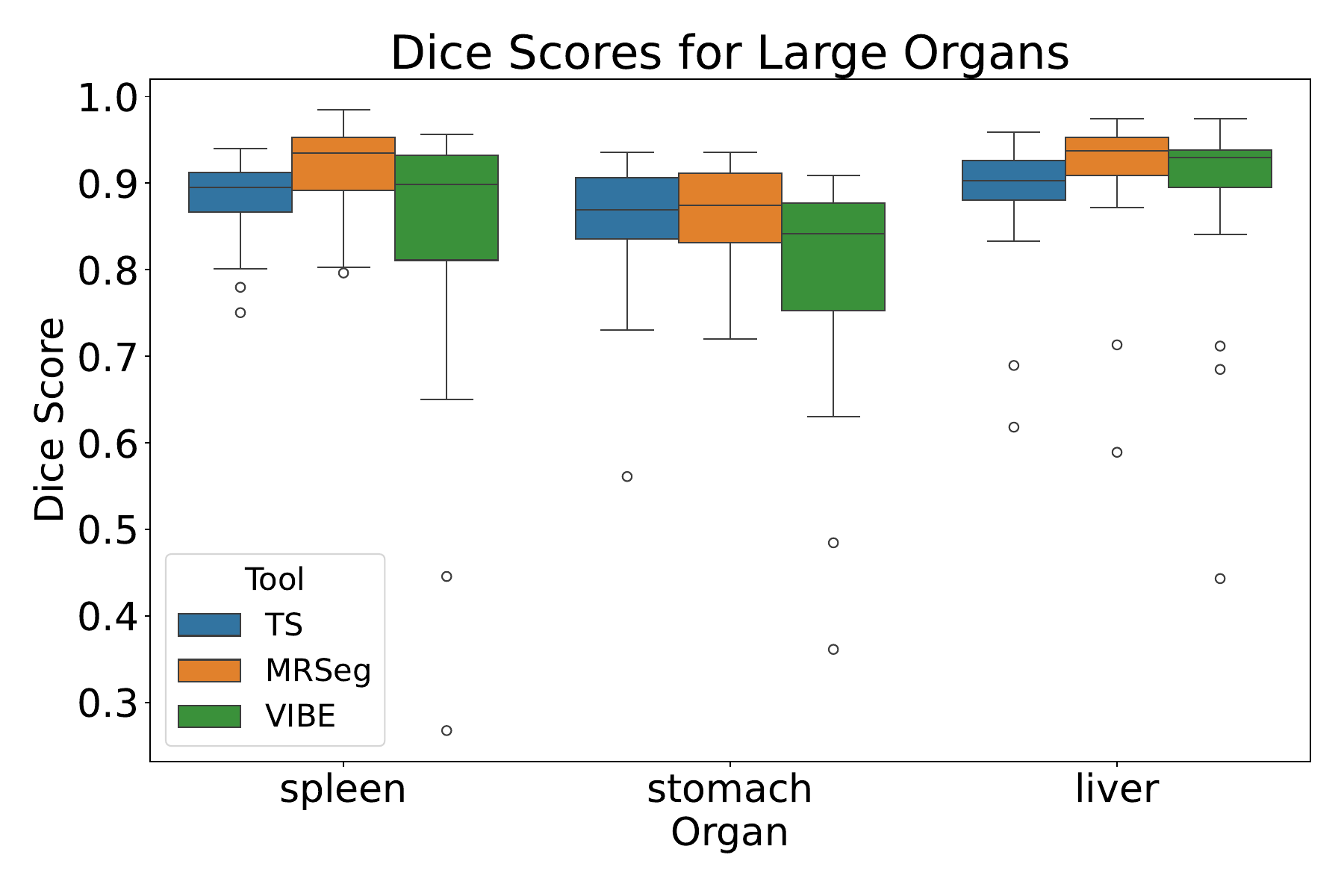}
\caption{Box plot comparing DSC of large abdominal organs (spleen, stomach, liver)}
\label{fig_DSC_largeOrgans}
\end{figure*}
%%%%%%%%%%%%%%%%%%%%%%%%%%%%%%%%%%%%%%%%%%%%%%%%%%%%%%%%%%%%%%%%%%

%%%%%%%%%%%%%%%%%%%%%%%%%%%%%%%%%%%%%%%%%%%%%%%%%%%%%%%%%%%%%%%%%%
\begin{figure*}[!htb]
% \begin{minipage}[b]{0.95\columnwidth}
%   \centering
%   \centerline{}
% %  \vspace{2.0cm}
%   \centerline
% \end{minipage}
\centering
\includegraphics[width=0.85\textwidth]{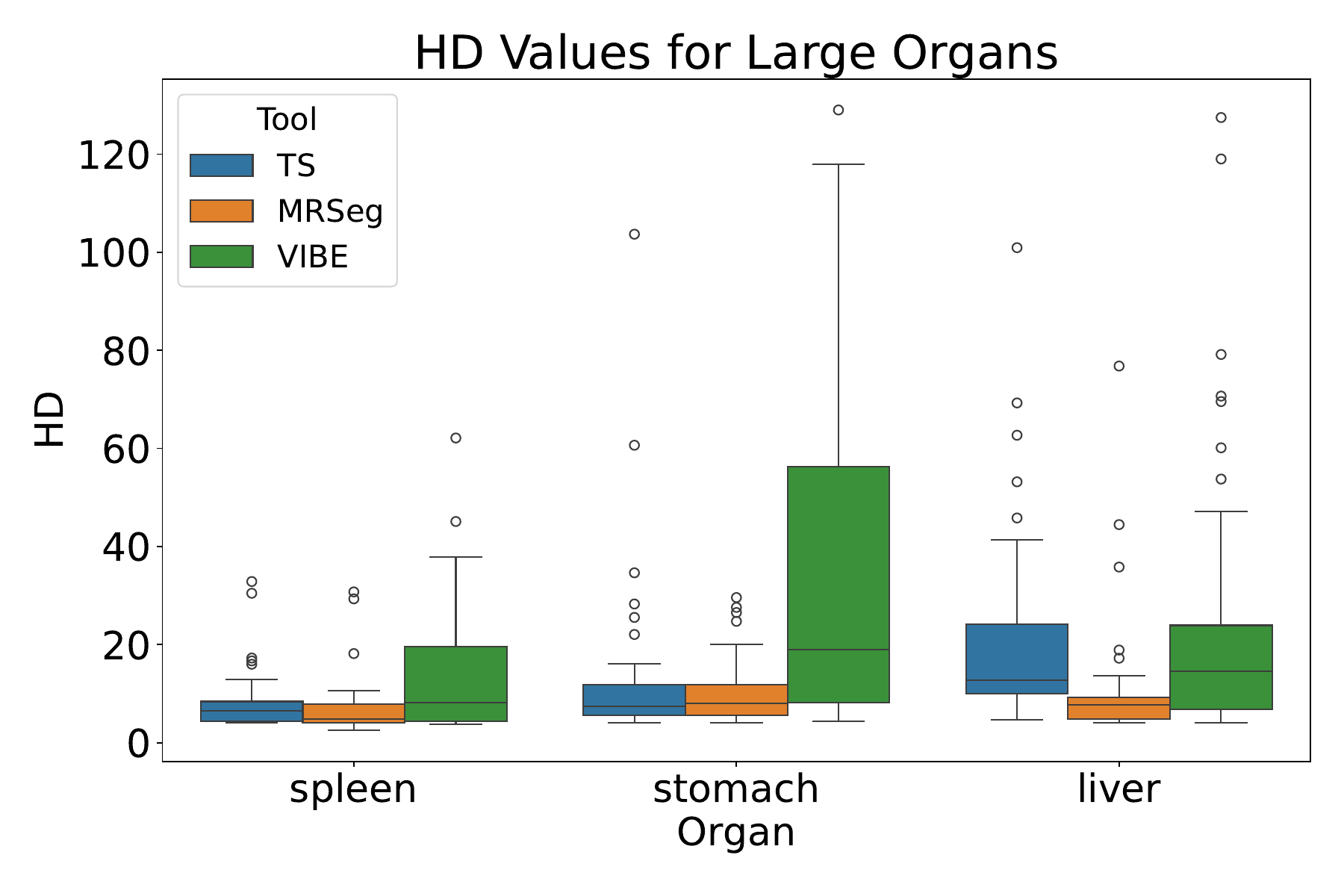}
\caption{Box plot comparing Hausdorff distances in mm of large abdominal organs (spleen, stomach, liver)}
\label{fig_hd_largeOrgans}
\end{figure*}
%%%%%%%%%%%%%%%%%%%%%%%%%%%%%%%%%%%%%%%%%%%%%%%%%%%%%%%%%%%%%%%%%%

%%%%%%%%%%%%%%%%%%%%%%%%%%%%%%%%%%%%%%%%%%%%%%%%%%%%%%%%%%%%%%%%%%
\begin{figure*}[!htb]
% \begin{minipage}[b]{0.95\columnwidth}
%   \centering
%   \centerline{}
% %  \vspace{2.0cm}
%   \centerline
% \end{minipage}
\centering
\includegraphics[width=0.85\textwidth]{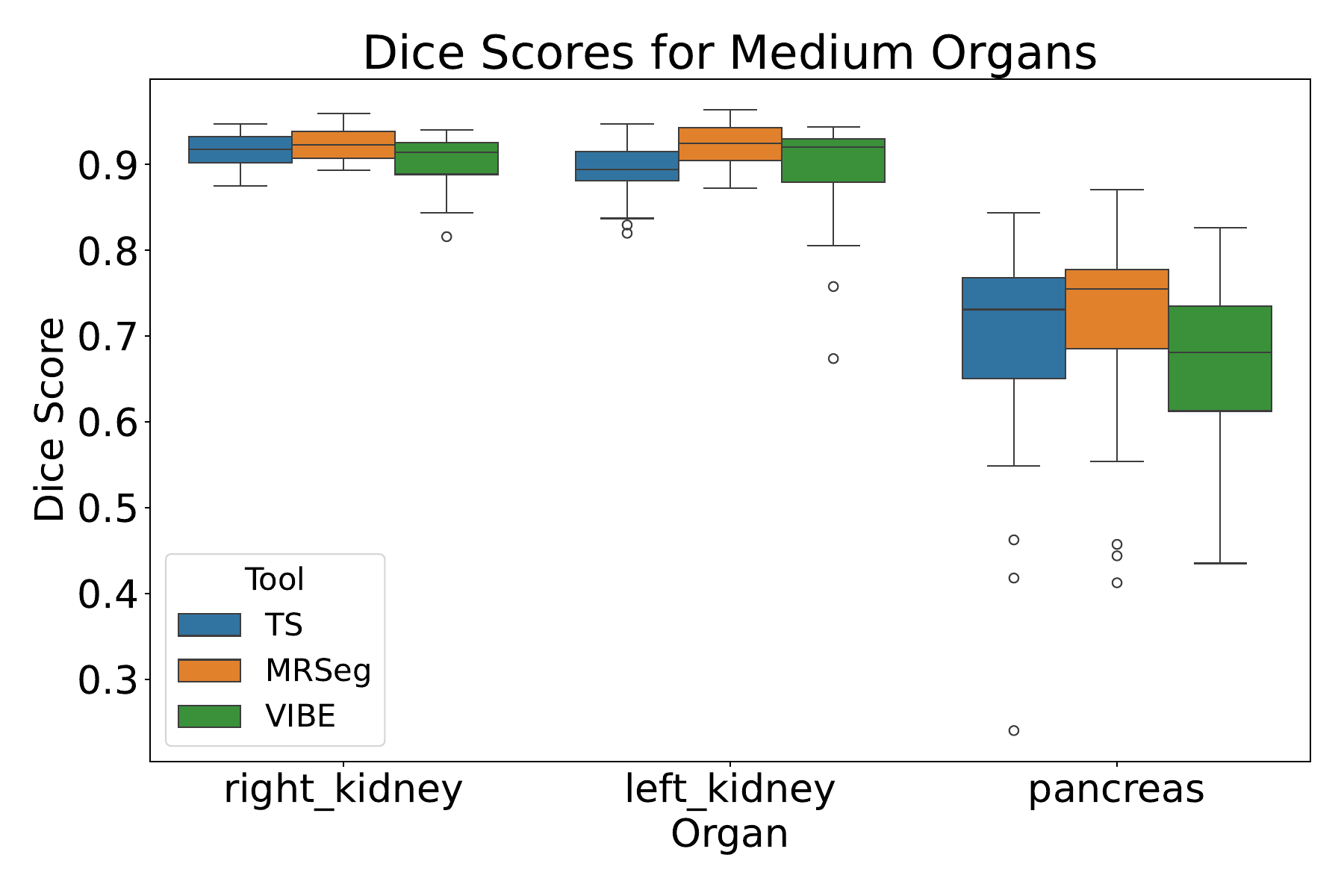}
\caption{Box plot comparing DSC of medium abdominal organs (right kidney, left kidney, pancreas)}
\label{fig_DSC_medOrgans}
\end{figure*}
%%%%%%%%%%%%%%%%%%%%%%%%%%%%%%%%%%%%%%%%%%%%%%%%%%%%%%%%%%%%%%%%%%

%%%%%%%%%%%%%%%%%%%%%%%%%%%%%%%%%%%%%%%%%%%%%%%%%%%%%%%%%%%%%%%%%%
\begin{figure*}[!htb]
% \begin{minipage}[b]{0.95\columnwidth}
%   \centering
%   \centerline{}
% %  \vspace{2.0cm}
%   \centerline
% \end{minipage}
\centering
\includegraphics[width=0.85\textwidth]{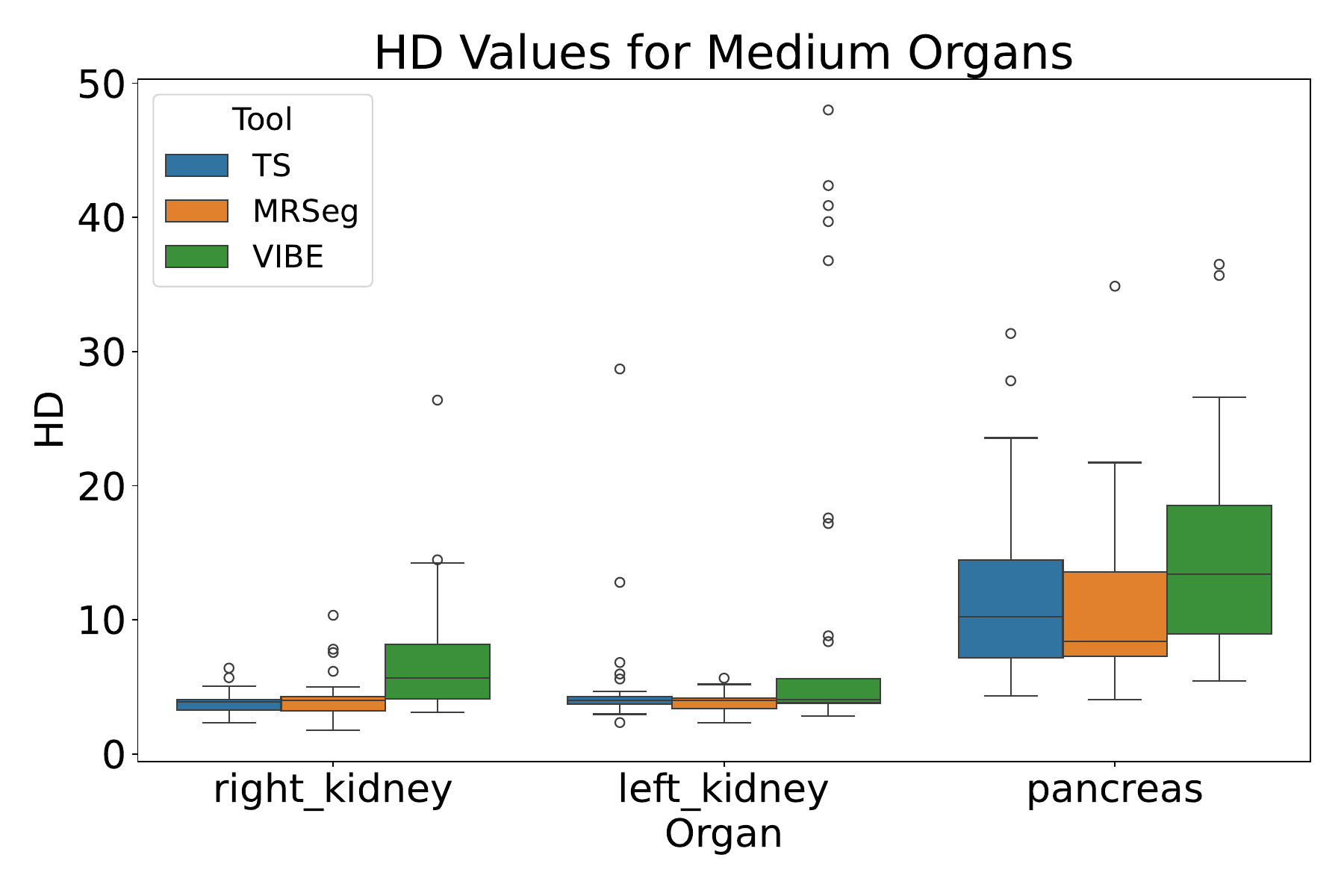}
\caption{Box plot comparing Hausdorff distances of medium abdominal organs (right kidney, left kidney, pancreas)}
\label{fig_hd_medOrgans}
\end{figure*}
%%%%%%%%%%%%%%%%%%%%%%%%%%%%%%%%%%%%%%%%%%%%%%%%%%%%%%%%%%%%%%%%%%

%%%%%%%%%%%%%%%%%%%%%%%%%%%%%%%%%%%%%%%%%%%%%%%%%%%%%%%%%%%%%%%%%%
\begin{figure*}[!htb]
% \begin{minipage}[b]{0.95\columnwidth}
%   \centering
%   \centerline{}
% %  \vspace{2.0cm}
%   \centerline
% \end{minipage}
\centering
\includegraphics[width=0.85\textwidth]{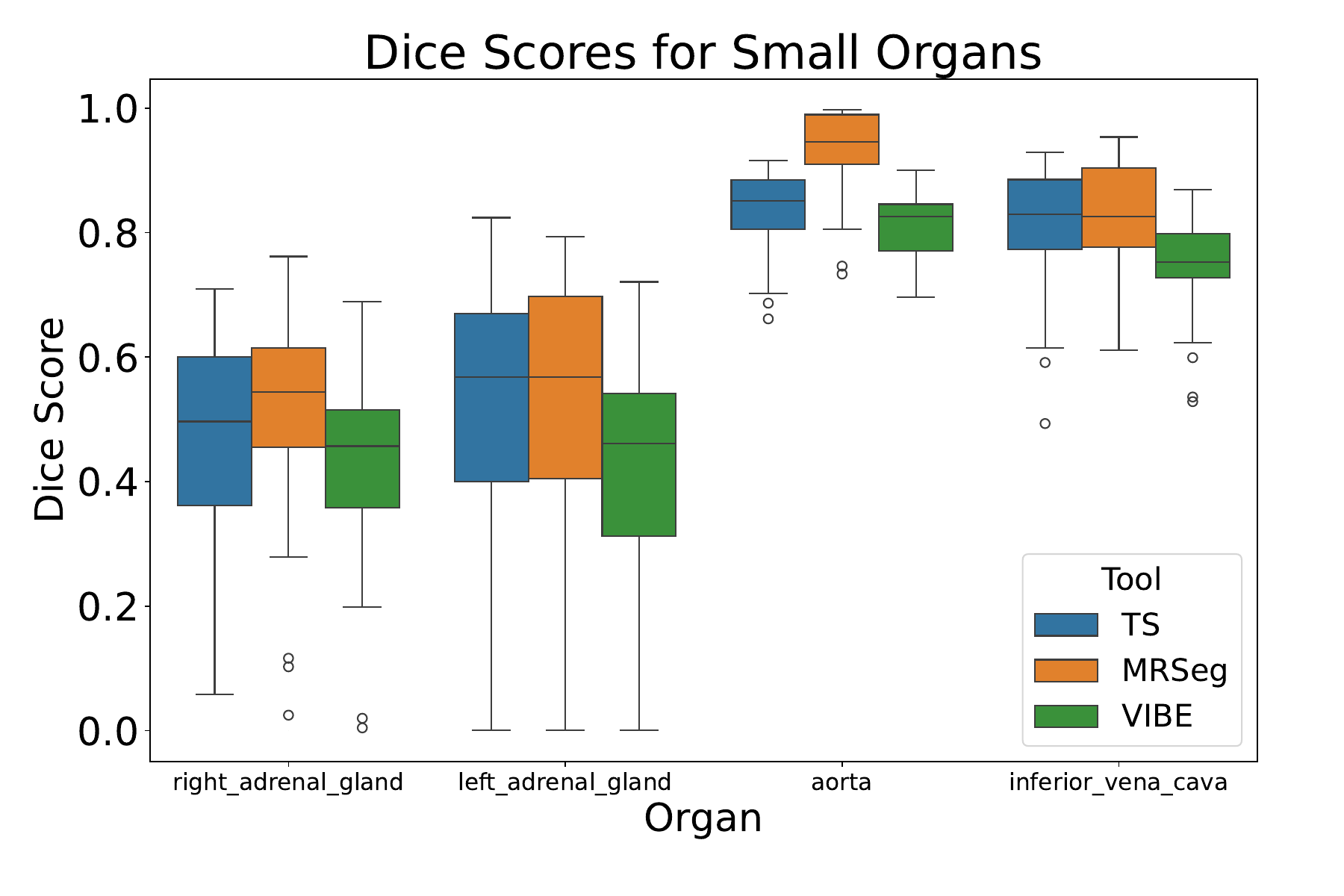}
\caption{Box plot comparing DSC of small abdominal organs (right adrenal gland, left adrenal gland, aorta, inferior vena cava)}
\label{fig_DSC_smallOrgans}
\end{figure*}
%%%%%%%%%%%%%%%%%%%%%%%%%%%%%%%%%%%%%%%%%%%%%%%%%%%%%%%%%%%%%%%%%%

%%%%%%%%%%%%%%%%%%%%%%%%%%%%%%%%%%%%%%%%%%%%%%%%%%%%%%%%%%%%%%%%%%
\begin{figure*}[!htb]
% \begin{minipage}[b]{0.95\columnwidth}
%   \centering
%   \centerline{}
% %  \vspace{2.0cm}
%   \centerline
% \end{minipage}
\centering
\includegraphics[width=0.85\textwidth]{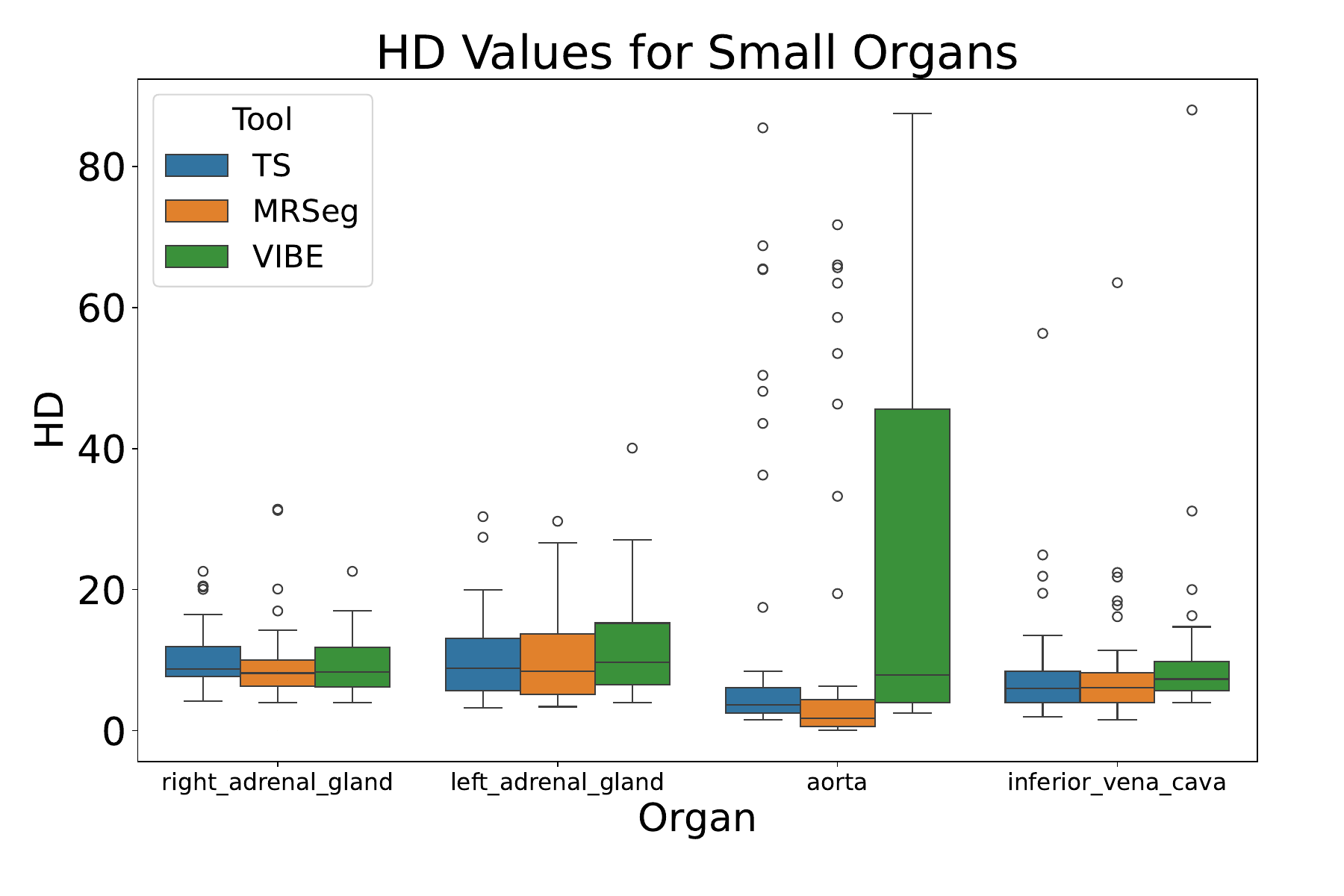}
\caption{Box plot comparing Hausdorff distances of small abdominal organs (right adrenal gland, left adrenal gland, aorta, inferior vena cava)}
\label{fig_hd_smallOrgans}
\end{figure*}
%%%%%%%%%%%%%%%%%%%%%%%%%%%%%%%%%%%%%%%%%%%%%%%%%%%%%%%%%%%%%%%%%%

%%%%%%%%%%%%%%%%%%%%%%%%%%%%%%%%%%%%%%%%%%%%%%%%%%%%%%%%%%%%%%%%%%
\begin{figure*}[!htb]
% \begin{minipage}[b]{0.95\columnwidth}
%   \centering
%   \centerline{}
% %  \vspace{2.0cm}
%   \centerline
% \end{minipage}
\centering
\includegraphics[width=\textwidth]{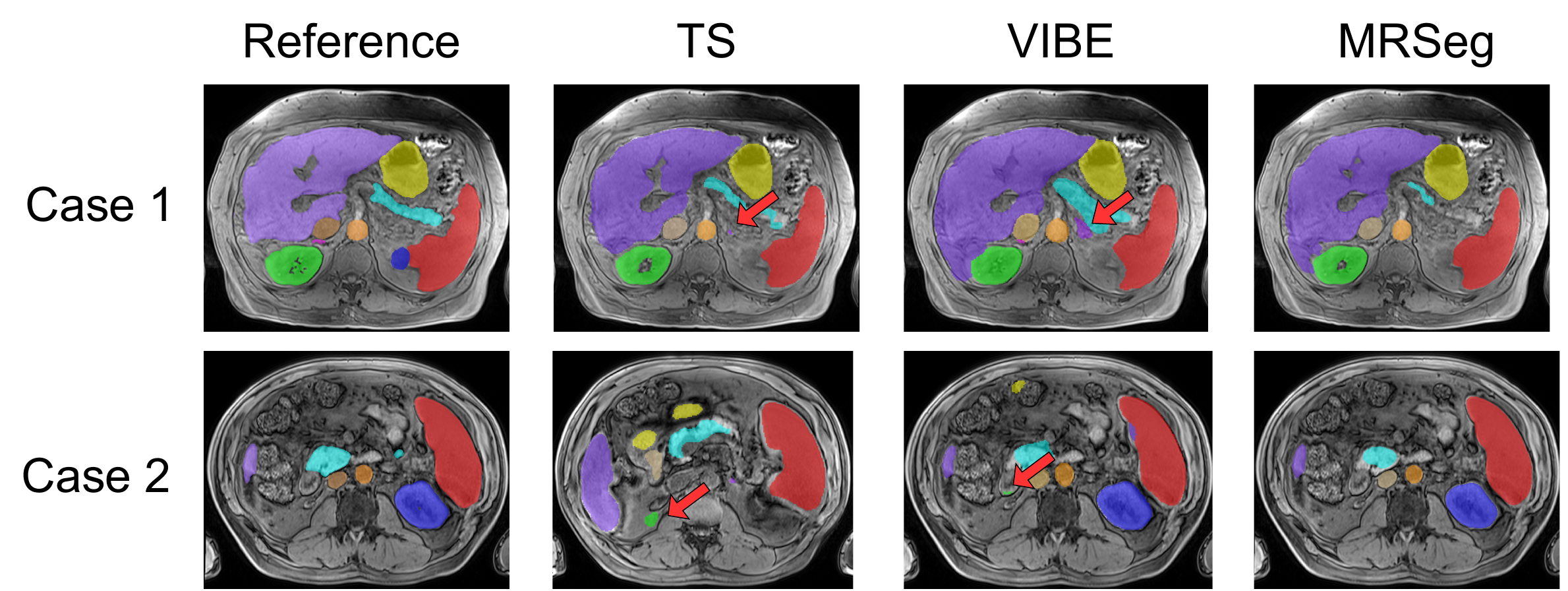}
\caption{False positive segmentations for the left adrenal gland (top row, red arrows) and right kidney (bottom row, red arrows) generated by TotalSegmentator MRI (TS) and TotalVibeSegmentator (VIBE). MRSegmentator did not generate any false positives on either case.}
\label{fig_falsePositiveSegmentations}
\end{figure*}
%%%%%%%%%%%%%%%%%%%%%%%%%%%%%%%%%%%%%%%%%%%%%%%%%%%%%%%%%%%%%%%%%%

\end{document}